%% file: main.tex
\documentclass[10pt,twocolumn,letterpaper]{article}

\usepackage[pagenumbers]{iccv} 


\definecolor{iccvblue}{rgb}{0.21,0.49,0.74}
\usepackage[pagebackref,breaklinks,colorlinks,allcolors=iccvblue]{hyperref}

\def\paperID{11451} 
\def\confName{ICCV}
\def\confYear{2025}

\title{Scoring, Remember, and Reference: Catching Camouflaged Objects in Videos}
\author{
	Yu'ang Feng\textsuperscript{\rm 1}, Shuyong Gao\textsuperscript{\rm 1,2,*},  Fuzhen Yan\textsuperscript{\rm 1},\\
    Yicheng Song\textsuperscript{\rm 1}, Lingyi Hong\textsuperscript{\rm 1},  Junjie Hu\textsuperscript{\rm 1}, Wenqiang Zhang\textsuperscript{\rm 1,*} \\ 
    \textsuperscript{\rm 1} Fudan University, Shanghai, China \\
    \textsuperscript{\rm 2} Keenon Robotics Co. Ltd, Shanghai, China
    }

\usepackage{multirow}
\usepackage{amsmath}

\begin{document}
\maketitle

\renewcommand{\thefootnote}{\fnsymbol{footnote}} 
\footnotetext[1]{Corresponding authors.}

\begin{abstract}

Video Camouflaged Object Detection (VCOD) aims to segment objects whose appearances closely resemble their surroundings, posing a challenging and emerging task. Existing vision models often struggle in such scenarios due to the indistinguishable appearance of camouflaged objects and the insufficient exploitation of dynamic information in videos. To address these challenges, we propose an end-to-end VCOD framework inspired by human memory-recognition, which leverages historical video information by integrating memory reference frames for camouflaged sequence processing. Specifically, we design a dual-purpose decoder that simultaneously generates predicted masks and scores, enabling reference frame selection based on scores while introducing auxiliary supervision to enhance feature extraction. Furthermore, this study introduces a novel reference-guided multilevel asymmetric attention mechanism, effectively integrating long-term reference information with short-term motion cues for comprehensive feature extraction. By combining these modules, we develop the \textbf{Scoring, Remember, and Reference (SRR)} framework, which efficiently extracts information to locate targets and employs memory guidance to improve subsequent processing. With its optimized module design and effective utilization of video data, our model achieves significant performance improvements, surpassing existing approaches by 10\% on benchmark datasets while requiring fewer parameters (54M) and only a single pass through the video. The code will be made publicly available.

\end{abstract}

\vspace{-0.2cm}
\begin{figure}
    \centering
    \includegraphics[width=1\linewidth]{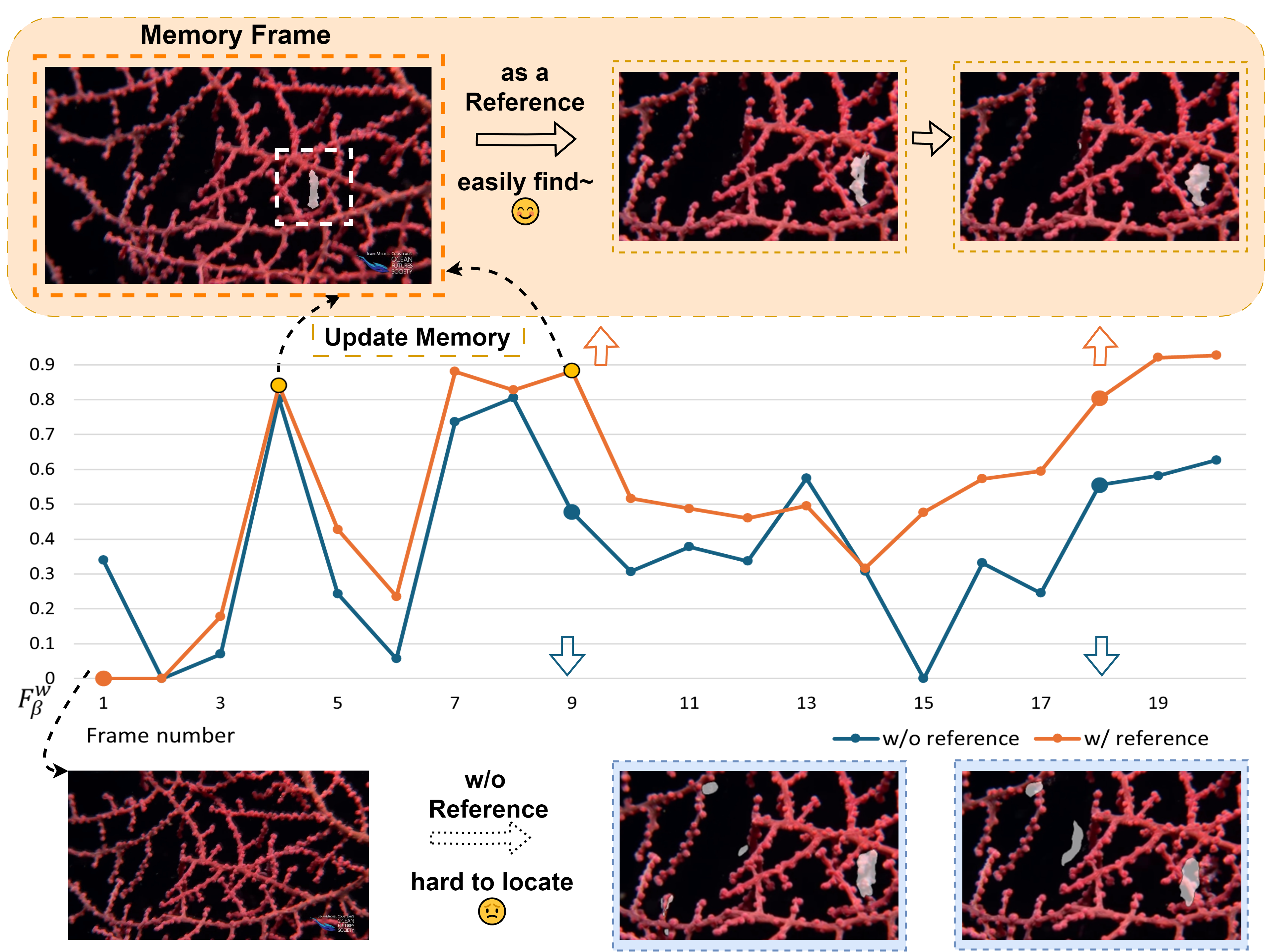}
    \caption{Comparison of methods with and without reference frame utilization. 
    Bottom: Conventional video models that do not save reference frames. 
    Top: Our method saves optimal frames as references to guide subsequent segmentation. 
    The line graph illustrates the performance curves of the two methods on a sample video, with bold points highlighting frames visualized.}
    \label{fig:figure1}
    \vspace{-0.3cm}
\end{figure}

\section{Introduction} 
When objects are concealed within backgrounds that closely resemble their appearance, even the human eye struggles to accurately identify them and delineate their boundaries. Camouflaged Object Detection (COD) is a task aimed at segmenting camouflaged objects from their surroundings. In real-world scenarios, numerous objects exhibit substantial visual similarity to their environment. This task finds applications in diverse fields, including wildlife conservation \cite{Perez2012-SpeciesDiscovery}, medical image segmentation \cite{fan2020inf,FanDP2020-Pranet,wu2021jcs}, art analysis \cite{Ge2018-Art,chu2010camouflage}, agriculture \cite{Rustia2020-Agriculture,liu2019pestnet}, and security \cite{LinCJ2019-Metaheuristic}. 

In dynamic environments, objects may be well-camouflaged while stationary, but human vision often identifies them readily upon movement. This highlights the critical role of dynamic information in object detection. Consequently, the task of Video Camouflaged Object Detection (VCOD) arises, focusing on identifying objects concealed within backgrounds from video sequences. Unlike typical video tasks, such as Video Object Segmentation (VOS) \cite{hong2023simulflow, miao2024region, yang2022decoupling} and Video Salient Object Detection (VSOD) \cite{lan2022siamese, gao2022weakly, ji2021full}, camouflaged targets are characterized by colors and textures that closely mimic the background and by indistinct boundaries. This necessitates the use of information beyond visual appearance.

Recent advancements in COD based on static images have achieved remarkable progress \cite{sun2022boundary,zhang2023predictive,huang2023feature}, offering valuable insights into leveraging visual information for this task. Nevertheless, methods relying solely on static information exhibit suboptimal performance when extended to video-based tasks. On the VCOD front, some progress has been achieved \cite{cheng2022implicit,hui2024implicit} with approaches that integrate motion information alongside appearance features. However, these methods often involve additional computational overhead and fail to fully exploit long-term memory information. 

The objective of this research is to propose methods that effectively utilize video data to extract dynamic information, thereby improving object detection and segmentation performance. How do humans identify an object concealed within its environment? Initially, the object may remain indistinguishable. However, since objects are rarely perfectly stationary, their movement or diminished camouflage enables humans to detect them by synthesizing new dynamic information with visual cues. More importantly, once identified, humans tend to remember the object's characteristics. Even if the object becomes stationary or re-hidden, its identification subsequently requires less effort due to memory. 

Inspired by this behavior, we propose the \textbf{Scoring, Remember, and Reference (SRR)} framework, named \textbf{SRR-Net}, which effectively utilizes video data to enhance VCOD performance. First, we introduce the concept of memory reference frames as the core of our design. Unlike ground-truth references provided for the first frame in VOS \cite{perazzi2016benchmark}, our reference frame is generated from preceding frames in the video sequence. Specifically, we develop a \textbf{Dual-Purpose Decoder} that outputs both segmentation maps and scores, enabling the model to select frames to retain as references. This design also incorporates automatic supervision for the scoring task, enhancing the quality of feature extraction.

Second, to utilize reference frames effectively, we propose \textbf{Reference-Guided Multilevel Asymmetric (RMA) Attention}. This attention mechanism within the transformer's backbone integrates memory information from reference frames with motion information from adjacent frames, enabling robust feature extraction for detecting camouflaged objects. Notably, our approach requires only a single pass through the video sequence for feature extraction and memory mechanism utilization, eliminating the need for future frames or additional post-processing, and enhancing its practicality for real-world applications.

In summary, the key contributions of this study are as follows:
\begin{itemize}
    \item We design a novel pipeline introducing the memory reference mechanism, effectively leveraging dynamic video information to advance the VCOD task and offering new methodologies for utilizing video data in VCOD.
    \item We propose the \textbf{RMA Attention} and the \textbf{Dual-Purpose Decoder}, implementing effective strategies for utilizing memory and motion information. These modules complete the \textbf{Scoring, Remember, and Reference} detection pipeline, enabling efficient detection of camouflaged objects in videos.
    \item Our model surpasses existing state-of-the-art methods by a margin of 10\% on large-scale datasets, requiring only a single pass through the video and fewer parameters (54M).
\end{itemize}

\begin{figure*}[h]
    \centering
    \includegraphics[width=1.0\linewidth]{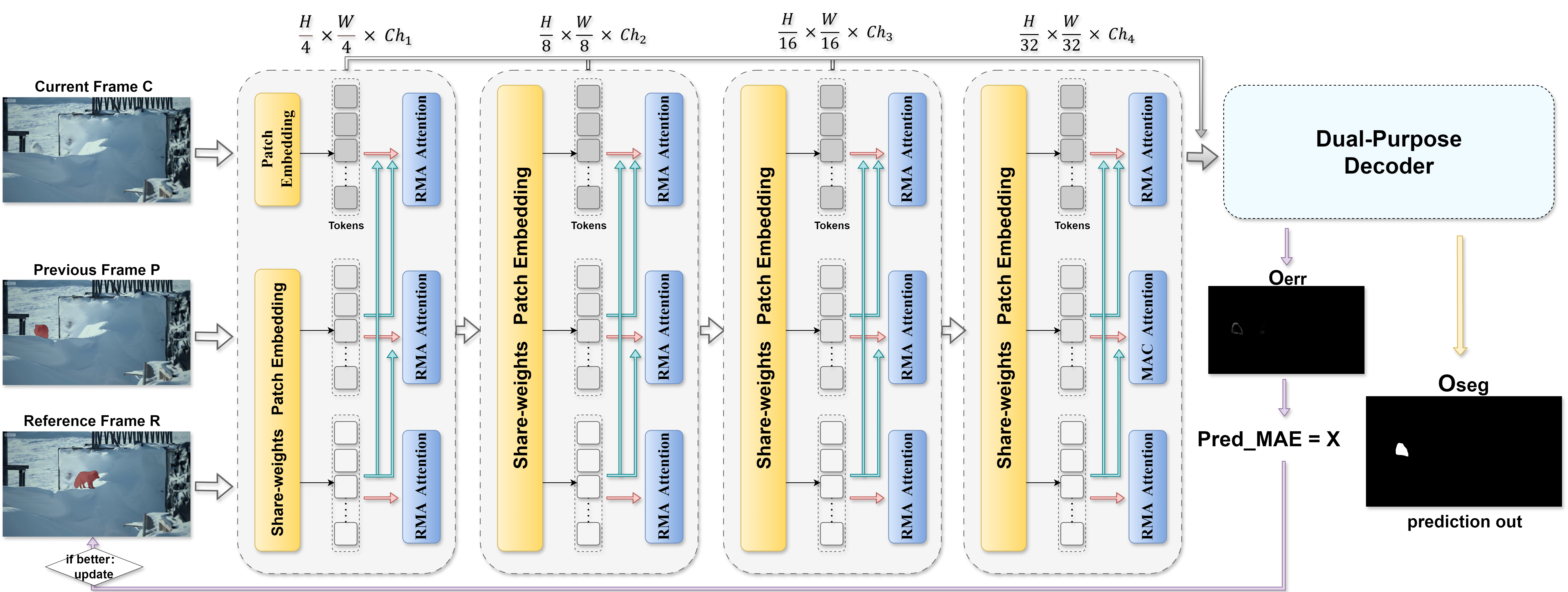}
    \caption{Overview of SRRNet. Using the proposed RMA attention mechanism, the framework constructs four stages to progressively extract multilevel features. These features are then fed into the dual-purpose decoder to generate segmentation results and predicted scores, which are subsequently used to update the reference frame.}
    \label{fig:figure2}

\vspace{-0.3cm}
\end{figure*}

\section{Related Work}

\subsection{Camouflaged Object Detection}
Camouflaged Object Detection (COD) focuses on identifying and segmenting objects concealed within their background. With the introduction of datasets such as \textbf{CAMO} \cite{camo}, \textbf{COD10K} \cite{fan2020camouflaged}, and \textbf{NC4K} \cite{Lv2021-RankNet}, increasing attention has been drawn to COD research.
Fan et al. \cite{fan2020camouflaged} proposed a two-stage strategy employing a coarse-to-fine approach to locate and segment camouflaged objects. Pang et al. \cite{pang2022zoom} mimicked human behavior by scaling input images to enhance target recognition. BGNet \cite{sun2022boundary} introduced edge-guided feature extraction to improve segmentation performance. SAM-COD \cite{chen2024sam} utilized the pretrained SAM model \cite{kirillov2023segment}, combined with weakly supervised labels, to perform COD. These studies have made significant contributions to COD. However, due to the high visual similarity between camouflaged objects and their background, relying solely on static visual information is insufficient for practical applications, leading to the emergence of video-based tasks for camouflaged object detection.

\subsection{Video Object Segmentation}
Video Object Segmentation (VOS) aims to segment desired objects from video sequences. The most common VOS task is Semi-Supervised VOS (SVOS) \cite{xu2018youtube}, where the ground-truth for the first frame is provided, and the model uses it as a reference to segment objects in subsequent frames \cite{yang2022decoupling,bekuzarov2023xmem++}. In contrast, the VCOD task does not provide a reliable segmentation for the first frame. Instead, in our framework, the concept of a reference frame is dynamically constructed during the processing of video sequences.

Another related task is Unsupervised Video Object Segmentation (UVOS), where no ground-truth is provided for the first frame. Hong et al.\cite{hong2023simulflow} tackled this task by combining motion information derived from optical flow with visual features for segmentation. Lee et al. \cite{lee2022iteratively} proposed a simple ResNet-based approach to select frames and transform UVOS into a standard SVOS problem. However, the ResNet module struggled with the complex visual features of camouflaged objects, and directly applying SVOS models proved unsuitable for VCOD. This has motivated the exploration of novel methods to introduce reference frames into camouflaged object detection.
\subsection{Video Camouflaged Object Detection}
Video Camouflaged Object Detection (VCOD) emerged to address the limitations of COD methods that rely solely on static image information. The objective of VCOD is to identify and segment objects hidden within the background from video sequences. Typically, VCOD tasks take consecutive frames of a video sequence as input and output binary segmentation maps for each frame.
In 2016, Pia Bideau et al. \cite{bideau2016s} introduced the first small-scale VCOD dataset, CAD. In 2020, Hala Lamdouar et al. \cite{lamdouar2020betrayed} proposed MoCA, a video dataset with bounding box annotations for camouflaged objects. Building upon this, Cheng et al.  \cite{cheng2022implicit} re-annotated and organized the data, presenting the first large-scale segmentation dataset for VCOD.Cheng et al. proposed SLT-Net \cite{cheng2022implicit}, which employs implicit motion extraction and uses sequence-to-sequence models to incorporate long-term information. Hui et al. \cite{hui2024implicit} introduced IMEX, integrating implicit and explicit motion learning into a unified framework to identify object motion. TSP-SAM \cite{hui2024endow} applied SAM to VCOD, leveraging its extensively pretrained feature extractor to provide prompts and inject spatiotemporal information, achieving promising results.
However, these methods involve additional processing. SLT-Net uses a two-step inference process to obtain long-term information. IMEX requires prior computation of optical flow fields before segmentation. TSP-SAM relies on SAM's feature extractor pretrained on the SA-1B dataset \cite{kirillov2023segment}. In comparison, our framework tightly integrates modules in an end-to-end manner, utilizing both short-term and long-term spatiotemporal information without external dependencies. This approach achieves superior performance while being more suitable for real-world real-time applications.

\section{Method} 

To enhance the utilization of video sequences for camouflaged object detection, we propose the framework illustrated in Figure \ref{fig:figure2}. The framework consists of two core components: the \textbf{Reference-Guided Multilevel Asymmetric (RMA) Attention Module} and the \textbf{Dual-Purpose Decoder}. The RMA module integrates auxiliary information from the previous frame and the reference frame (along with its mask) to achieve multilevel attention-based feature extraction. To determine which frames to store as memory, we designed a decoder that simultaneously performs segmentation and score prediction by introducing a scoring branch. Combining these modules, we establish a complete segmentation-selection-memory-utilization pipeline. Built on a multilevel backbone inspired by PVT \cite{wang2021pyramid}, we construct the \textbf{Scoring, Remember, and Reference Network (SRRNet)}. This pipeline operates using only sequential RGB frame inputs from video sequences, without additional preprocessing, and outputs corresponding foreground-background segmentation maps.

\begin{figure}
    \centering
    \includegraphics[width=1\linewidth]{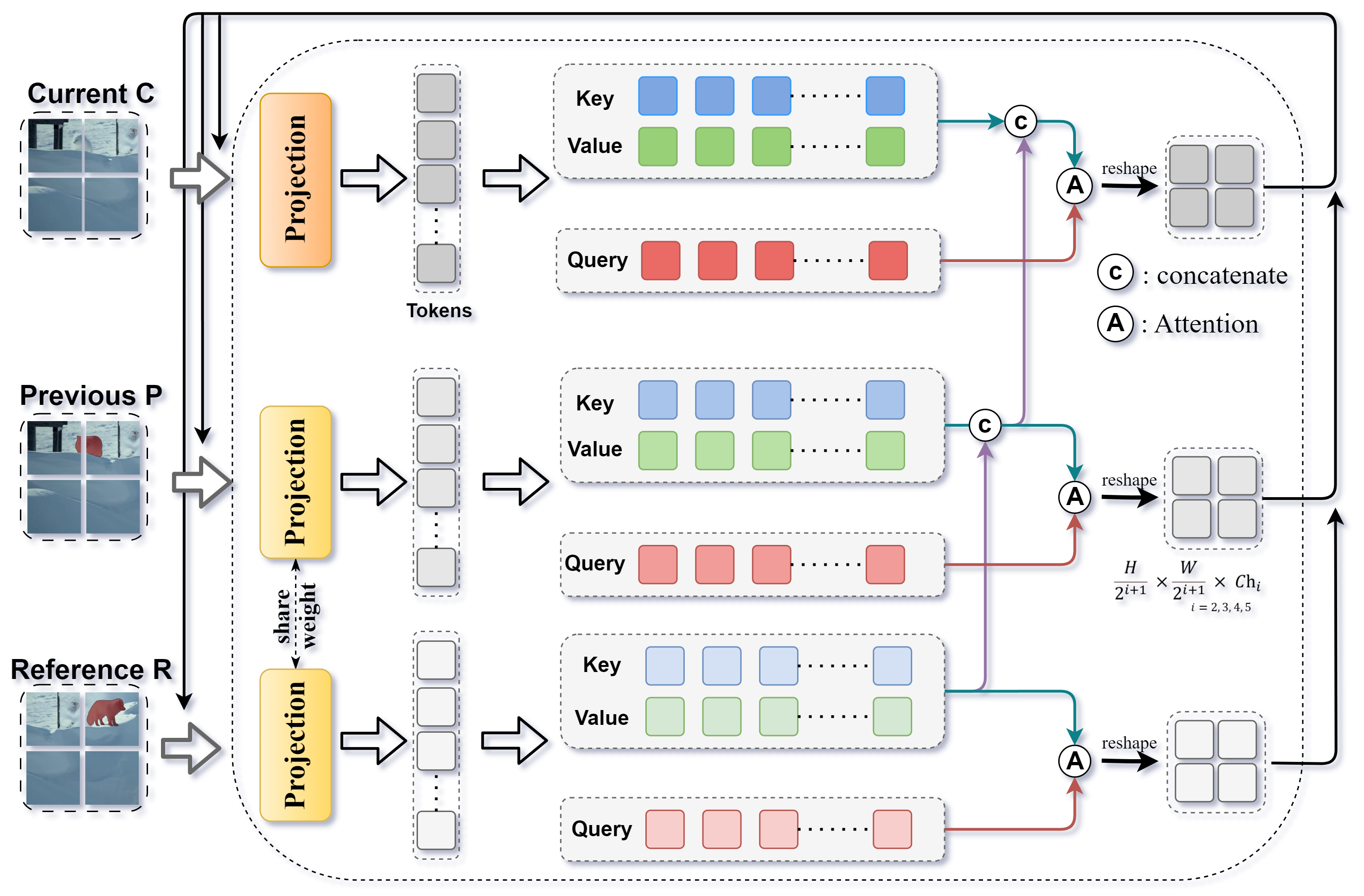}
    \caption{The RMA attention mechanism consists of self-attention and cross-attention. The cross-attention operates asymmetrically across the three branches.}
    \label{fig:figure3}
    \vspace{-0.3cm}
\end{figure}
\subsection{Reference-Guided Multilevel Asymmetric (RMA) Attention Module}

In VCOD tasks, the high similarity between objects and their background makes segmentation challenging when relying solely on appearance information. To leverage the dynamic nature of video data, we introduce implicit motion information and memory information with a novel asymmetric attention mechanism. The \textit{previous frame} implicitly provides temporal context for the \textit{current frame}. As shown in SLT-NET \cite{cheng2022implicit}, implicit motion features offer advantages in VCOD tasks, as optical flow can mislead models. Our approach eliminates the need for additional video traversal to compute optical flow \cite{teed2020raft}, allowing segmentation to be completed in a single pass. Motion information from the previous frame is complemented by the reference frame for feature extraction.

Inspired by SimulFlow \cite{hong2023simulflow}, which employs cross-attention between optical flow and image features, we extend this concept by introducing a reference frame branch. As shown in Figure \ref{fig:figure3}, the module inputs include the current frame $C_{\text{img}}$, the processed \textit{previous frame} $P_{\text{img}}$ with its predicted mask $P_{\text{msk}}$, and the reference frame $R_{\text{img}}$ with its stored mask $R_{\text{msk}}$. Unlike VOS tasks that use first-frame ground-truth, VCOD reference frames are dynamically selected and stored during video processing. The inputs are tokenized via patch embedding, producing $C$, $P$, and $R$, which undergo self-attention for feature extraction. Multilevel asymmetric cross-attention is then applied: $R$ attends unidirectionally to both $C$ and $P$, while $P$ attends unidirectionally to $C$. The attention operation is formulated as:
\begin{align}
    \text{Attention}(Q, K, V) = \text{softmax}\left(\frac{QK^T}{\sqrt{d}}\right)V,
\end{align}
where $Q$, $K$, $V$ denote query, key, and value embeddings, and $d$ is the dimension of $K$. Feature maps are processed as follows:
\begin{equation}
\begin{aligned}
    K_u &= \text{Concat}(K_P, K_R), & V_u &= \text{Concat}(V_P, V_R), \\
    K_w &= \text{Concat}(K_C, K_P, K_R), & V_w &= \text{Concat}(V_C, V_P, V_R).
\end{aligned}
\end{equation}
The resulting attention maps are:
\vspace{-6pt} 
\begin{equation}
\begin{aligned}
    A_R &= \text{Attention}(Q_R, K_R, V_R), \\
    A_P &= \text{Attention}(Q_P, K_u, V_u), \\
    A_C &= \text{Attention}(Q_C, K_w, V_w),
\end{aligned}
\end{equation}
\vspace{-6pt} 

where $A_R$, $A_P$, and $A_C$ represent attention maps for the reference, previous, and current frames, respectively. Storing raw frames instead of feature maps enables flexible reprocessing for complex camouflage scenarios. Using unidirectional asymmetric attention avoids negative interactions and reduces computational costs, balancing performance and efficiency.

\subsection{RMA Transformer}

Based on the RMA module, we design a pyramid-structured transformer decoder inspired by PVT \cite{wang2021pyramid} and adapted from the Mix Transformer encoders (MiT) in SegFormer \cite{xie2021segformer}. As shown in Figure \ref{fig:figure2}, the transformer consists of four stages, progressively extracting multilevel features with overlapping patch embeddings and RMA attention. Inputs include the current frame $C_{\text{img}}$, the previous frame $P_{\text{img}}$ with its mask $P_{\text{msk}}$, and the reference frame $R_{\text{img}}$ with its mask $R_{\text{msk}}$, with input dimensions $H \times W \times 3$ for frames and $H \times W \times 1$ for masks. Concatenating $P_{\text{img}}$ and $P_{\text{msk}}$ (and similarly for $R$) yields $H \times W \times 4$ inputs. In the first stage, images are divided into $4 \times 4$ patches and tokenized, with shared weights for $P$ and $R$, while $C$ uses independent weights. Multilevel RMA attention generates feature maps, with $C_i \in \mathbb{R}^{H_i \times W_i \times \text{Ch}_i}$ representing the feature map for $C$ at stage $i$, where $H_i = H/2^{i+1}$, $W_i = W/2^{i+1}$, and $i \in \{1, 2, 3, 4\}$. Feature maps for $P_i$ and $R_i$ are similarly defined. These $4 \times 3$ feature maps are passed to the decoder for further processing.

\subsection{Dual-Purpose Decoder for Segmentation and Scoring} 

Maintaining a reliable reference frame is essential for effective guidance in processing video sequences. In semi-supervised video object segmentation tasks \cite{perazzi2016benchmark}, the first frame and its ground-truth provide a strong reference for subsequent segmentation. However, in VCOD, no ground-truth is available for the first frame. To fully utilize video sequences, we design a module capable of selecting the optimal reference frame alongside its predicted segmentation map. The proposed framework completes this task in a single traversal of the video, avoiding methods that rely on comparisons after processing the entire sequence \cite{lee2022iteratively}. Instead, a segmentation score is output in real-time, enabling improved reference frame selection by comparing stored scores.

\begin{figure}[ht]
    \centering
    \includegraphics[width=1\linewidth]{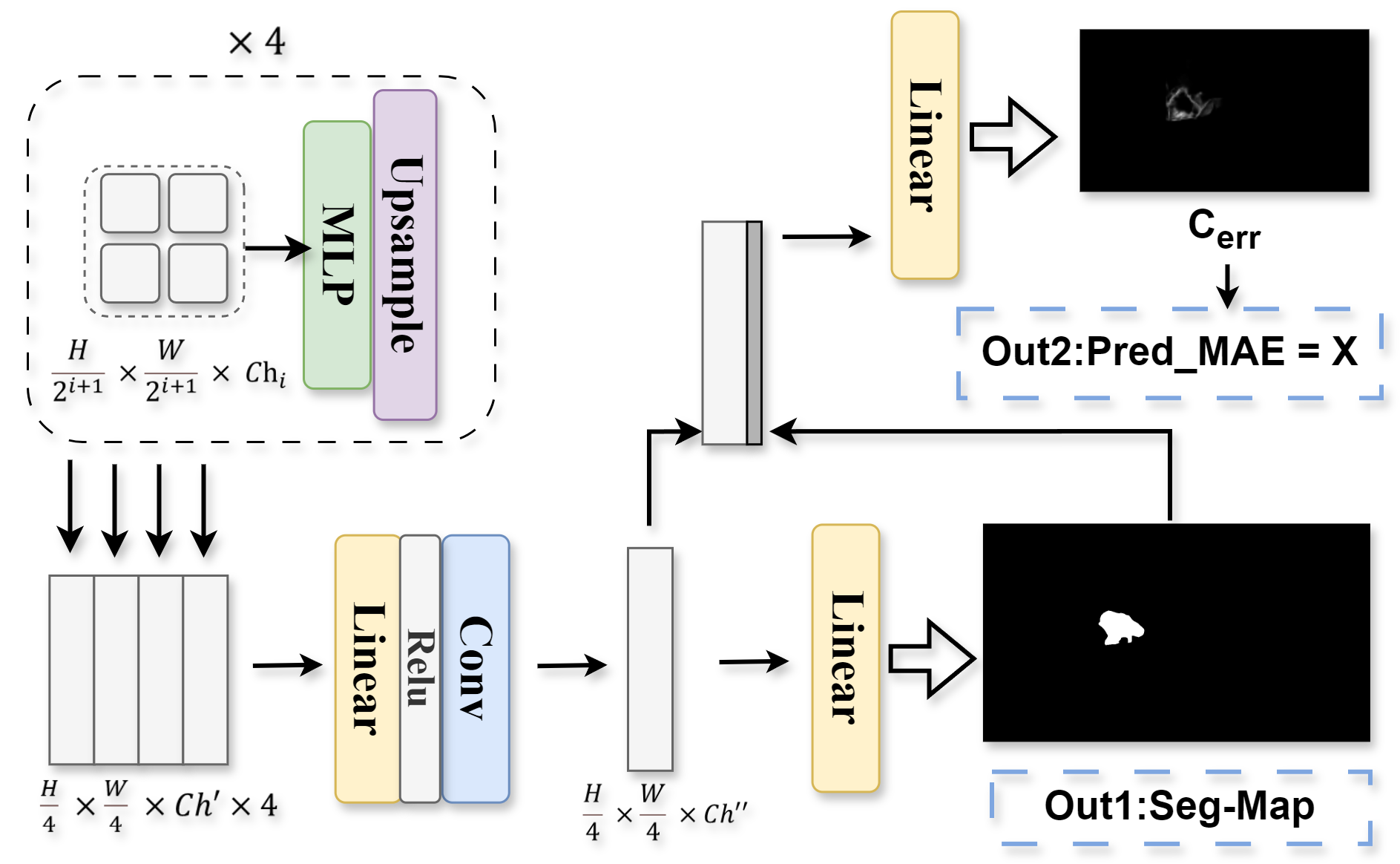}
    \caption{Dual-purpose decoder. The decoder outputs both a segmentation map and a predicted error map.}
    \label{fig:figure4}
    \vspace{-0.3cm}
\end{figure}

Given the challenges of VCOD, approaches like adding a lightweight network for score prediction \cite{lee2022iteratively} prove ineffective. To address this, we augment the decoder with an additional branch based on encoder-extracted features. A lightweight decoder integrates these features using simple MLPs, as illustrated in Figure \ref{fig:figure4}. After processing features from the transformer, the decoder outputs results through two branches: one generates a binary segmentation map $O_{\text{msk}}$, while the other predicts a segmentation score based on encoder features and the segmentation map.

Outputting a single score can hinder backpropagation and reduce robustness. To address this, we estimate the pixel-wise error between $O_{\text{msk}}$ and the ground-truth. By averaging, we derive the predicted Mean Absolute Error (MAE) of the segmentation map as the score. The predicted output $O_{\text{err}}$ is a score matrix of the same dimensions, enabling backpropagation during training. For features extracted at each encoder stage, $C_i, P_i, R_i \in \mathbb{R}^{H_i \times W_i \times \text{Ch}_i}$ $(i \in \{1, 2, 3, 4\})$, the following operations are applied:
\begin{equation}
\begin{aligned}
    F_i &= \text{Linear}(\text{Ch}_i, \text{Ch}') \, \text{Concat}(C_i, P_i, R_i), \\
    F_i &= \text{Upsample}(H/4 \times W/4)(F_i),
\end{aligned}
\end{equation}
where $\text{Linear}(\text{Ch}_{\text{in}}, \text{Ch}_{\text{out}})(X)$ projects $X$ from $\text{Ch}_{\text{in}}$ to $\text{Ch}_{\text{out}}$, and $\text{Upsample}(h \times w)(X)$ resizes $X$ to dimensions $h \times w$. Here, $\text{Ch}'$ is set to 1024.

The processed features are used for prediction:
\begin{equation}
\begin{array}{c}
    \displaystyle F = \text{Linear}(4*\text{Ch}', \text{Ch}') \, \text{Concat}(F_1, F_2, F_3, F_4), \\[3pt]
    \displaystyle F = \text{Conv}(\text{Ch}', \text{Ch}'')(F), \\[3pt]
    \displaystyle M = \text{Linear}(\text{Ch}'', 2)(F), \\[3pt]
    \displaystyle O_{\text{msk}} = \text{argmax}(\text{softmax}(M)) \in \{0, 1\}.
\end{array}
\end{equation}
where $\text{Conv}(\text{Ch}_{\text{in}}, \text{Ch}_{\text{out}})(X)$ represents a convolution operation with kernel size 3 and stride 1, mapping $X$ from $\text{Ch}_{\text{in}}$ to $\text{Ch}_{\text{out}}$, with $\text{Ch}''$ set to 256. Softmax and argmax generate the final mask indicating the foreground or background.

The second branch estimates the score:
\begin{equation}
\begin{aligned}
    F' &= \text{Concat}(F, M), \\
    O_{\text{err}} &= \text{Linear}(\text{Ch}'', 1)(F') \in (0, 1).
\end{aligned}
\end{equation}

The resulting score matrix $O_{\text{err}}$ approximates the pixel-wise absolute error between $O_{\text{msk}}$ and the ground-truth. By averaging, we obtain the predicted MAE score for $O_{\text{msk}}$, which evaluates frames for selecting suitable reference frames.

The score prediction branch provides a novel supervision mechanism that enhances feature extraction for evaluating segmentation quality. This supervision is self-generated during training, requiring no additional input. Further discussions on this mechanism are provided in Section \ref{sec:sec4.3}.

\noindent\textbf{Score Prediction Demonstration:}
As shown in Figure 5, several videos from the MoCA-Mask test set are used as examples. In the line graph, the horizontal axis represents the frame indices of the videos, while the vertical axis denotes the MAE values for each frame. The two curves correspond to the actual MAE values of the segmentation map and the predicted MAE values. It is evident from the line graph that the predicted scores generally align well with the actual MAE scores. While the model cannot fully fit frames with significant variations, the peaks and valleys of the predicted curve closely match those of the actual curve, indicating that the model effectively captures the trend of sequential error changes, which facilitates the selection of suitable reference frames based on the predicted scores.
﻿
The score prediction module outputs an error matrix $O_{\text{err}}$, which has the same size as the segmentation map and represents the estimated pixel-wise error between the segmentation map and the ground truth. Figure 5 also includes visualizations of real and predicted error maps for selected frames. These visualizations demonstrate that the model can reasonably produce deeper or shallower error maps based on the magnitude of segmentation errors, even identifying problematic areas in the predicted segmentation maps. This highlights the effectiveness of the score prediction module.

Moreover, the potential applications of the error map may extend further. We conducted preliminary experiments to utilize the error map for refining segmentation. However, due to the wide range and instability of values within the error map, we have yet to identify a robust approach for reliably improving segmentation quality through backpropagation. This represents a potential direction for future research. Nonetheless, the module's reliability in predicting error trends remains undeniable.

\subsection{Training and Inference} 

\noindent\textbf{Inference:}
The goal is to complete object detection and segmentation for the entire video sequence in a single traversal. The proposed framework processes video sequences strictly in sequential order. During inference, the only external input required is the current frame image, serving as the model's $C_{\text{img}}$ input. Inputs for the previous and reference frames are automatically stored and reused from the processed frames. 

After processing each frame, the model outputs the predicted mask $O_{\text{msk}}$ and the predicted error matrix $O_{\text{err}}$ for the current frame. The current frame's image and mask are saved for the next frame, and the predicted MAE score is computed as the average of all pixel values in $O_{\text{err}}$. This score is compared to the saved score $S$ from the memory module, and if the new score is smaller, the reference frame is updated to the recently processed frame, improving reference frame selection in real time.

For the first frame of a video, $P_{\text{img}}$ and $R_{\text{img}}$ are initialized to be identical to the first frame, while $P_{\text{msk}}$ and $R_{\text{msk}}$ are set to zeros, and the recorded score $S$ is initialized to 1. This inference process avoids second-pass traversal and does not utilize future frames while processing the current frame. Predicted masks for each frame are generated sequentially.

\noindent\textbf{Training:} 
To enhance parallel efficiency and ensure data diversity, sequential video traversal is not employed during training. Instead, frames are selected randomly. One frame is designated as the current frame $C_{\text{img}}$, with its preceding frame and ground-truth used as $P_{\text{img}}$ and $P_{\text{seg}}$. Additionally, a random frame earlier in the sequence is chosen as the reference frame $R_{\text{img}}$ and $R_{\text{seg}}$. This strategy enables efficient, parallelized training.

Following prior work \cite{cheng2022implicit}, pre-training was performed using static images from the COD10K dataset. For static images, neighboring images of the same category and random images were used as $P_{\text{img}}, P_{\text{seg}}, R_{\text{img}},$ and $R_{\text{seg}}$. The segmentation map $O_{\text{msk}}$, predicted by the model, was supervised using binary cross-entropy loss. 
For the predicted score matrix $O_{\text{err}}$, the real error matrix was calculated, and supervision was applied at each position using a mean squared error (MSE) loss.The real error matrix was computed by subtracting the predicted segmentation map $O_{\text{msk}}$ from the ground-truth, which then served as the target for $O_{\text{err}}$. During supervision $O_{\text{err}}$, the decoder branch responsible for $O_{\text{msk}}$ was frozen to prevent interference with segmentation performance. While the supervision of $O_{\text{err}}$ introduces additional loss terms, the required new ground-truth is generated automatically without external input.

The final loss function combines binary cross-entropy loss for segmentation and MSE loss for score prediction:
\begin{align}
    \mathcal{L} = \mathcal{L}_{\text{bce}}(O_{\text{msk}}, \text{gt}) + \gamma \mathcal{L}_{\text{MSE}}(O_{\text{err}}, \text{gt} - O_{\text{msk}}^*),
\end{align}
where $\mathcal{L}_{\text{bce}}$ and $\mathcal{L}_{\text{MSE}}$ denote the binary cross-entropy and mean squared error losses, respectively. $\text{gt}$ represents the ground-truth segmentation map, $O_{\text{msk}}$ is the predicted segmentation mask, $O_{\text{msk}}^*$ is its detached copy, and $O_{\text{err}}$ is the predicted error matrix. The parameter $\gamma$ balances the contribution of the two loss terms.

\begin{figure}
    \centering
    \includegraphics[width=1\linewidth]{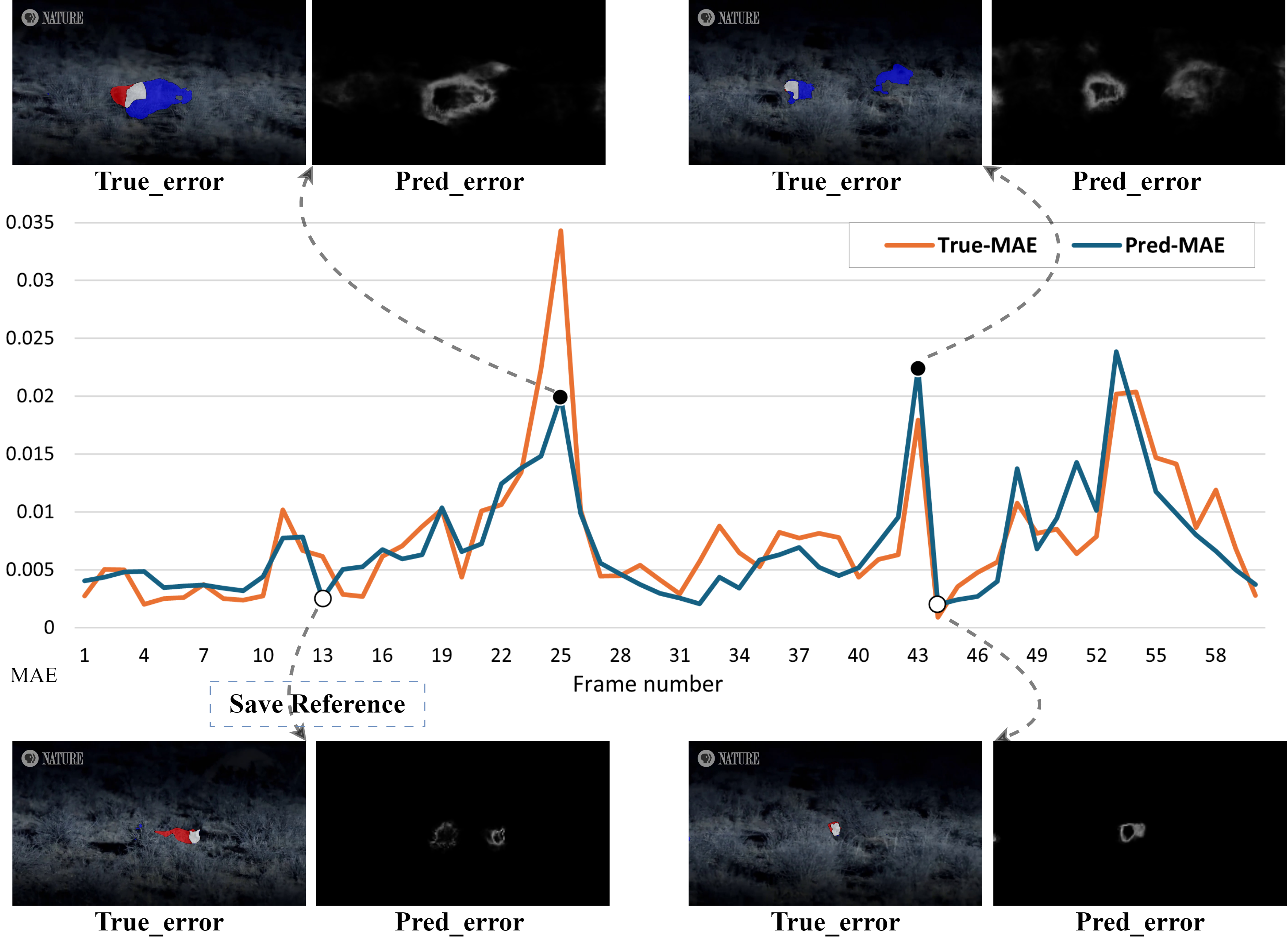}
    \caption{Comparison between true error and predicted error. 
    In each pair of images, the right image represents the predicted error, while the left image shows the segmentation results, where red areas indicate false negatives, blue areas represent false positives, and white areas denote correctly predicted regions. 
    The line graph compares the predicted MAE values with the actual MAE values, demonstrating that our prediction module effectively captures the trend of error variations between frames.
}
    \label{fig:figure5}
    \vspace{-0.3cm}
\end{figure}

\begin{table*}[!h]
    \centering
    \caption{Quantitative comparisons with state-of-the-art methods on MoCA-Mask and CAD. 
    "$\downarrow$" indicates smaller values are better, while "$\uparrow$" indicates larger values are better. 
    The best results are highlighted in \textbf{bold}. }
    \label{tab:table1}

    \resizebox{\textwidth}{!}{
    \begin{tabular}{c c c ccccc c ccccc}
        \toprule
         \multirow{2}{*}{\textbf{Model}} & \multirow{2}{*}{\textbf{Input}} & \multirow{2}{*}{\textbf{Params}} & \multicolumn{5}{c}{\textbf{MoCA-Mask}} & \multirow{2}{*}{} & \multicolumn{5}{c}{\textbf{CAD}} \\
        \cline{4-8} \cline{10-14}
           & & & $\mathbf{S_\alpha} \uparrow$ & $\mathbf{F^{w}_{\beta}} \uparrow$ & $\mathbf{M} \downarrow$ & \textbf{mDice} $\uparrow$ & \textbf{mIoU} $\uparrow$ & & $\mathbf{S_\alpha} \uparrow$ & $\mathbf{F^{w}_{\beta}} \uparrow$ & $\mathbf{M} \downarrow$ & \textbf{mDice} $\uparrow$ & \textbf{mIoU} $\uparrow$ \\
        \cline{1-8} \cline{10-14}
         SINet(CVPR'20)\cite{fan2020camouflaged} &Image & 49.05M & 0.598 & 0.231 & 0.028 & 0.276 & 0.202 & & 0.636 & 0.346 & 0.041 & 0.381 & 0.283 \\
         SINet-V2(TPAMI'21)\cite{FanDP2021-ConcealedOD} &Image & 27.04M & 0.588 & 0.204 & 0.031 & 0.245 & 0.180 & & 0.653 & 0.382 & 0.039 & 0.413 & 0.318 \\
         ZoomNet(CVPR'22)\cite{pang2022zoom} &Image & 32.38M & 0.582 & 0.211 & 0.033 & 0.224 & 0.167 & & 0.633 & 0.349 & 0.033 & 0.349 & 0.273 \\
         DGNet(MIR'23)\cite{ji2023deep} &Image & 21.02M & 0.571 & 0.175 & 0.035 & 0.222 & 0.156 & & 0.686 & 0.416 & 0.037 & 0.456 & 0.340 \\
         FSPNet(CVPR'23)\cite{huang2023feature} &Image & 274.24M & 0.594 & 0.182 & 0.044 & 0.238 & 0.167 & & 0.681 & 0.401 & {0.044} & 0.446 & 0.332 \\ 
         FEDER(CVPR'23)\cite{He2023-FEDER} &Image & 44.20M & 0.560 & 0.165 & 0.033 & 0.194 & 0.137 & & 0.691 & 0.444 & {0.029} & 0.474 & 0.375 \\

        \cline{1-8} \cline{10-14}
        RCRNet(ICCV'19)\cite{yan2019semi} &Video & 53.85M & 0.555 & 0.138 & 0.033 & 0.171 & 0.116 & & 0.627 & 0.287 & 0.048 & 0.309 & 0.229 \\
         PNS-Net (MICCAI'21)\cite{ji2021progressively} &Video & 26.94M & 0.544 & 0.097 & 0.033 & 0.121 & 0.101 & & 0.655 & 0.325 & 0.048 & 0.384 & 0.290 \\
         MG (ICCV'21)\cite{yang2021self} &Video & 4.77M & 0.530 & 0.168 & 0.067 & 0.181 & 0.127 & & 0.594 & 0.336 & 0.059 & 0.368 & 0.268 \\
         SLT-Net (CVPR'22)\cite{cheng2022implicit} &Video & 82.38M & 0.634 & 0.317 & 0.027 & 0.356 & 0.271 & & {0.696} & {0.471} & 0.031 & 0.480 & {0.392} \\
         SLT-Net-Long (CVPR'22)\cite{cheng2022implicit} &Video & 82.31M & {0.631} & {0.311} & {0.026} & {0.367} & {0.279} & & {0.691} & {0.481} & {0.030} & {0.493} &{0.401} \\
         IMEX(TMM'24)\cite{hui2024implicit}  &Video & 32.16M & {0.661}  & {0.371} & {0.020} &{0.409} & {0.319} & &{0.695} & {0.490} & {0.030} & {0.501} & {0.412} \\
         TSP-SAM-Point(CVPR'24)\cite{hui2024endow}  &Video & 89.78M* & {0.673}  & {0.400} & {0.012} &{0.421} & {0.345} & &{0.681} & {0.500} & {0.031} & {0.496} & {0.393} \\
         TSP-SAM-Bbox(CVPR'24)\cite{hui2024endow}  &Video & 89.78M* & {0.689}  & {0.444} & \textbf{0.008} &{0.458} & {0.388} & &{0.704} & {0.524} & {0.028} & {0.543} & {0.438} \\
         \rowcolor{gray!30} \textbf{Ours(SRRNet)}  &Video & 53.79M & \textbf{0.727} &\textbf{0.489} & \textbf{0.008} &\textbf{0.513} & \textbf{0.428} & & \textbf{0.723} & \textbf{0.541} & \textbf{0.025} & \textbf{0.553} & \textbf{0.452} \\
        \bottomrule
    \end{tabular}
    }\vspace{-0.1cm}
\end{table*}

\section{Experiments}

\subsection{Experimental Setup}

In this section, we thoroughly validate the performance of the proposed framework on two existing VCOD datasets.

\noindent\textbf{Datasets:} 
\textbf{MoCA-Mask} \cite{cheng2022implicit} is currently the only large-scale segmentation dataset in this field. It is based on the Moving Camouflaged Animals (MoCA) Dataset \cite{lamdouar2020betrayed}, with selected data annotated with masks. This dataset contains 87 video sequences of camouflaged animals and provides 5,750 mask annotations. The \textbf{Camouflaged Animal Dataset (CAD)} \cite{bideau2016s} is an earlier, smaller VCOD dataset that consists of only 9 short clips of camouflaged animals sourced from YouTube videos. In this study, CAD is used solely for validation purposes.

\noindent\textbf{Evaluation Metrics.} 
We employ five evaluation metrics for validation in this section. \textbf{S-measure} ($S_{\alpha}$) \cite{fan2017structure} evaluates structural similarity. \textbf{Weighted F-measure} ($F_{\omega\beta}$) \cite{margolin2014evaluate} combines precision and recall for comprehensive evaluation. \textbf{Mean Absolute Error (MAE)} measures pixel-level absolute differences, while \textbf{Average Dice (mDice)} and \textbf{Average IoU (mIoU)}  assess the segmentation quality of both foreground and background regions.

\noindent\textbf{Implementation Details.} 
During training, all input images are first resized to maintain their original aspect ratios, with the shorter edge scaled to 512 pixels. Random cropping is then applied to produce $512 \times 512$ patches, which are further augmented through random flipping, photometric distortion, and normalization before being input into the model for training.

The model is trained in parallel on three RTX 3090 GPUs, with a total batch size of 12. The AdamW optimizer \cite{Loshchilov2017DecoupledWD} is used for optimization. Following previous studies, a two-stage training strategy is adopted. In the first stage, the model is pretrained on the COD10K dataset using static images. In the second stage, the model is fine-tuned on the MoCA-Mask dataset. During pretraining, the learning rate is set to $6 \times 10^{-5}$ and reduced by half after 50,000 iterations. For fine-tuning, a learning rate of $1 \times 10^{-5}$ is used, with a maximum of 10,000 training iterations.

\subsection{Baseline Evaluation}

\noindent\textbf{Baseline:} 
To evaluate our model, we compared it against several state-of-the-art (SOTA) methods. The performances of these methods, as shown in Table 1, are derived from publicly available code, predicted masks provided by the authors, or results reported in their respective papers. Some methods are image-based COD models, including SINet \cite{fan2020camouflaged}, SINet-v2 \cite{FanDP2021-ConcealedOD}, ZoomNet \cite{pang2022zoom}, DGNet \cite{ji2023deep}, FEDER \cite{He2023-FEDER}, and FSPNet \cite{huang2023feature}, which detect and segment camouflaged objects based on static image appearances.

Other methods are VCOD models designed for video sequences, such as RCRNet \cite{yan2019semi}, PNS-Net \cite{ji2021progressively}, MG \cite{yang2021self}, SLT-Net \cite{cheng2022implicit}, IMEX\cite{hui2024implicit}, and TSP-SAM \cite{hui2024endow}. Among these, RCRNet, PNS-Net, and MG were not explicitly developed for camouflaged object detection, with MG being a self-supervised approach relying solely on optical flow for supervision.

Following prior experimental setups \cite{cheng2022implicit}, these methods are pretrained on the COD10K dataset and fine-tuned on the MoCA-Mask dataset. Certain video-based methods involve multiple passes over video sequences; for example, SLTNet-Long \cite{cheng2022implicit} obtains long-term results by performing additional inference on the outputs of its short-term model. IMEX requires prior computation of optical flow fields before segmentation, while TSP-SAM leverages a SAM pretrained feature extractor on the SA-1B dataset \cite{kirillov2023segment}. In contrast, our method processes video sequences in a single pass without any additional preprocessing, yet it outperforms all other methods.

\begin{figure*}
    \centering
    \includegraphics[width=1\linewidth]{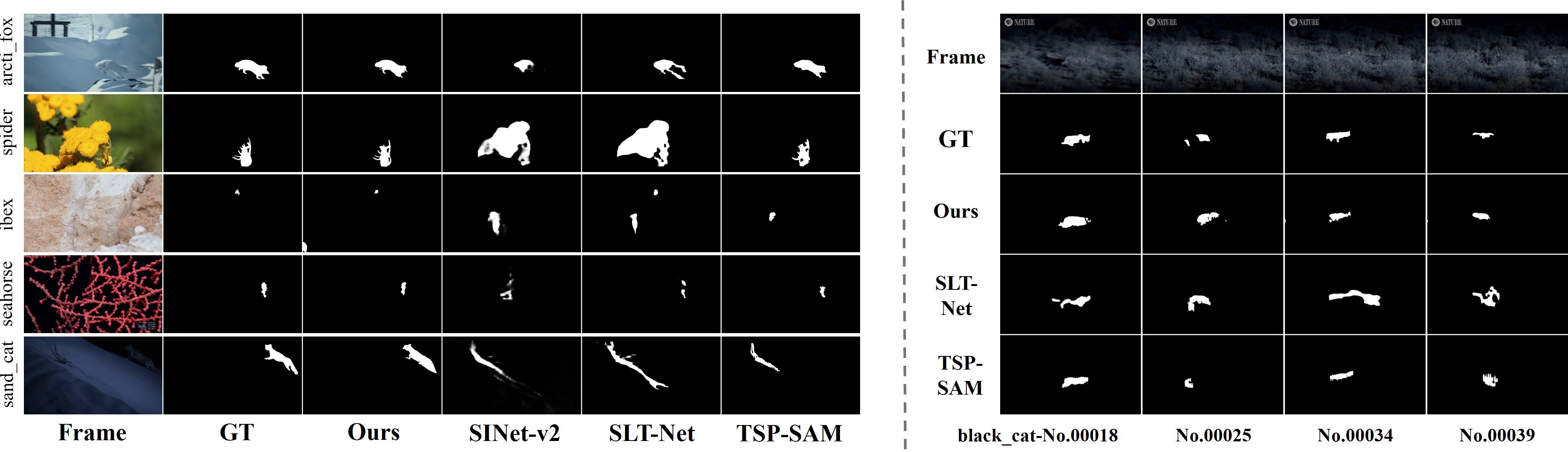}
    \caption{Qualitative comparison on the MoCA-Mask dataset. 
    The left side presents a horizontal comparison across five video sequences, while the right side illustrates comparisons between different frames within the same video sequence.}
    \label{fig:figure6}
\end{figure*}

\textbf{Quantitative Results.} 
Table 1 presents a quantitative comparison between the proposed SRRNet and other SOTA methods on the MoCA-Mask and CAD datasets. The results demonstrate that our model outperforms all existing methods across all evaluation metrics. Notably, video-based VCOD methods consistently surpass static image-based methods, indicating that video introduces additional information that significantly enhances segmentation performance. By effectively leveraging video data, SRRNet achieves remarkable performance improvements on the MoCA-Mask dataset. Compared to the current best methods, SRRNet delivers over 10\% gains in $F^{w}_{\beta}$, mDice, and mIoU metrics.
On the CAD dataset, our method also outperforms prior approaches, achieving improvements of 2\% to 12\%. The relatively modest gains on CAD may stem from its shorter video lengths, where SRRNet's memory and reference mechanisms are less impactful. Despite these performance improvements, SRRNet comprises only 53.79M parameters, significantly fewer than SLT-Net (82.3M) and TSP-SAM (89.78M*, fine-tuned components only). Combined with its single-pass design, SRRNet demonstrates substantial advantages in real-time detection and segmentation for practical applications.
\noindent\textbf{Qualitative Comparison:} 
To provide a more intuitive comparison of segmentation performance, we selected six videos from the MoCA-Mask dataset as examples and qualitatively compared SRRNet with other SOTA methods. As shown in Figure \ref{fig:figure6}, our method successfully identifies hidden objects (e.g., video ``ibex''), maintains stable performance (e.g., video ``black\_cat''), and achieves more detailed segmentation results (e.g., video ``spider'').

\subsection{Ablation Study}\label{sec:sec4.3}

To evaluate the effectiveness of the proposed modules, we performed ablation experiments on the MoCA-Mask dataset:

\noindent\textbf{Scoring, Memory, and Reference Mechanism:} 
We designed a scoring, memory, and reference process to select and utilize reference frames. To validate its efficacy, we conducted quantitative experiments summarized in Table 3, with performance improvements illustrated in Figure \ref{fig:figure1}:
\textbf{Row 1:} No reference mechanism. The model uses two consecutive frames as input, with feature extraction via two branches, resembling standard video models \cite{cui2022mixformer}, achieving basic VCOD performance.
 \textbf{Row 2:} Adds a reference frame branch using the proposed RMA attention module for feature extraction, without decoder modifications. This results in an average 20\% performance improvement.
\textbf{Row 3:} Introduces a dual-purpose decoder with an error prediction branch supervised during training. Inference does not utilize scores, and reference frames are selected randomly, showing further performance improvement.
\textbf{Row 4:} The full model, incorporating all mechanisms, achieves the highest performance among all configurations.

\begin{table}[h]
    \centering
    \caption{Ablation study for the SRRNet mechanism. 
    w/o: Without the module; w: With the module. 
    }
    \label{tab:table2}
    \resizebox{0.47\textwidth}{!}{
    \begin{tabular}{c ccccc}
        \toprule
         \textbf{Model} & $\mathbf{S_\alpha}$ $\uparrow$ & $\mathbf{F^{w}_{\beta}}$ $\uparrow$ & $\mathbf{M}$ $\downarrow$ & \textbf{mDice} $\uparrow$ & \textbf{mIoU} $\uparrow$ \\
        \midrule
         w/o reference & 0.684 & 0.398 & 0.012 & 0.419 & 0.352 \\
         w/ RMA transformer& {0.705} & {0.454} & {0.008} & 0.470 & {0.391} \\
         w/ DP decoder & {0.713} & {0.476} & \textbf{0.007} & 0.486 & {0.415} \\
         \textbf{RMA+DP+score(Ours)} & \textbf{0.727} & \textbf{0.489} & {0.008} & \textbf{0.513} & \textbf{0.428} \\
        \bottomrule
    \end{tabular}
    }
\end{table}

\noindent\textbf{RMA Attention:}
To assess the proposed RMA attention module, which employs multilevel asymmetric cross-attention, we conducted experiments detailed in Table 2:
\textbf{Row 1:} Only self-attention is applied within the three feature extraction branches.
\textbf{Row 2:} Single-level cross-attention from the previous frame to the current frame is introduced.
\textbf{Row 3:} Full bidirectional cross-attention across all branches is used, but incurs significant computational costs.
\textbf{Row 4:} The proposed asymmetric cross-attention from the reference and previous frames to the current frame achieves the best balance of performance and efficiency.
The results demonstrate that the RMA attention mechanism effectively balances computational efficiency and segmentation performance compared to alternative designs.

\vspace{-0.1cm}
\begin{table}[!h]
    \centering
    \caption{Ablation study for the RMA attention mechanism, evaluating different designs of cross-attention.
    }
    \label{tab:table3}
    \vspace{-0.1cm}
    \resizebox{0.47\textwidth}{!}{
    \begin{tabular}{c ccccc}
        \toprule
         \textbf{Model} & $\mathbf{S_\alpha}$ $\uparrow$ & $\mathbf{F^{w}_{\beta}}$ $\uparrow$ & $\mathbf{M}$ $\downarrow$ & \textbf{mDice} $\uparrow$ & \textbf{mIoU} $\uparrow$ \\
        \midrule
         w/o cross Attn & 0.707 & 0.465 & 0.010 & 0.489 & 0.402 \\
         w/ motion Attn & {0.710} & {0.477} & {0.009} & 0.502 & {0.410} \\
         Full Attn & {0.716} & {0.483} & {0.009} & \textbf{0.522} & \textbf{0.428} \\
         \textbf{RMA Attn(Ours)} & \textbf{0.727} & \textbf{0.489} & \textbf{0.008} & {0.513} & \textbf{0.428} \\
        \bottomrule
    \end{tabular}
    }
    \vspace{-0.3cm}
\end{table}

\subsection{Limitations}

Despite the effectiveness of the proposed framework, several limitations remain. First, the model processes video frames strictly in sequential order. While this design is advantageous for real-time applications, it restricts the ability to revise earlier predictions based on subsequent frames and partially limits parallel processing capabilities. Second, the current utilization of error maps for refining segmentation does not fully exploit their potential in guiding error correction, leaving room for improvement. Lastly, in scenarios prone to inaccuracies, error accumulation over the sequence may adversely affect performance.

\section{Conclusion}

Inspired by the memory-recognition behavior of humans in sequential tasks, this paper proposed a novel \textbf{Scoring, Remember, and Reference (SRR)} process for VCOD. The proposed framework incorporates a scoring module that autonomously selects reference frames, while leveraging a unique asymmetric attention design to efficiently extract features supported by spatiotemporal information. 

The end-to-end framework, referred to as \textbf{SRRNet}, introduces a new methodology for effectively utilizing spatiotemporal dynamics in VCOD tasks. Through extensive experiments, SRRNet demonstrated superior performance and computational efficiency compared to existing state-of-the-art approaches. These results underscore the potential of the SRR process and the broader applicability of the proposed design in advancing research on camouflaged object detection in videos.

{
    \small
    \bibliographystyle{ieeenat_fullname}
    \bibliography{main}
}

\end{document}